\documentclass[journal]{IEEEtran}
\usepackage{amsmath}
\usepackage{amsfonts}
\usepackage{algcompatible}
\usepackage{algorithm}
\usepackage{array}
\usepackage{textcomp}
\usepackage{stfloats}
\usepackage{url}
\usepackage{verbatim}
\usepackage{graphicx}
\usepackage{cite}
\usepackage{amssymb}
\usepackage{subcaption}
\usepackage[table]{xcolor}
\usepackage{multirow}
\usepackage{booktabs}
\usepackage{mathtools}
\usepackage{tablefootnote}

\newcommand{\ra}[1]{\renewcommand{\arraystretch}{#1}}

\begin{document}
	
	\title{An Event-based Algorithm for Simultaneous 6-DOF Camera Pose Tracking and Mapping}
	
	\author{Masoud Dayani Najafabadi, Mohammad Reza Ahmadzadeh %
		\thanks{The authors are with the Department of Electrical and Computer Engineering, Isfahan University of Technology, Isfahan, Iran.}}

	\maketitle
	
	\begin{abstract}
		Compared to regular cameras, Dynamic Vision Sensors or Event Cameras can output compact visual data based on a change in the intensity in each pixel location asynchronously. In this paper, we study the application of these novel sensors to current image-based SLAM techniques. To this end, the information in adaptively selected event windows is processed to form motion-compensated images. These images are then used to reconstruct the arbitrary scene and estimate the 6-DOF pose of the camera. To our knowledge, this is the first event-only algorithm that utilizes SLAM modules like local mapping to estimate the pose based on the reconstructed map and can use various distortion models for the camera. In addition, we propose an inertial version of the event-only pipeline to assess its capabilities. We compare the results of different configurations of the proposed algorithm against the ground truth for sequences of several publicly available event datasets. We also compare the results of the proposed pipeline with two representative methods and show it can produce comparable or more accurate results provided the map estimate is reliable.
	\end{abstract}
	
	
	\section{Introduction}
	In applications such as robot navigation and control, augmented reality, and 3D reconstruction, we need to know a device's location according to a map representation of the environment. Sometimes we are interested in estimating either the map or the robot's pose. Localization or odometry is the problem of estimating the robot's pose when the map is known. On the other hand, the general problem of estimating the device's pose and the scene structure in parallel is called simultaneous localization and mapping (SLAM).
	
	The main challenge in an odometry/SLAM system is to achieve accurate and robust performance in real-time, no matter how severe the conditions are. Certain circumstances, including poor lighting, fast motion, dynamic scene, and long outdoor distances, can adversely complicate the problem. This problem is challenging because the estimation generally requires the optimization of nonlinear equations with a large number of unknown parameters.
	
	One way to address this issue is to use algorithmic techniques and constructs such as parallelizing different parts of the system or using efficient data structures \cite{ptam_2007}, \cite{orb_slam_Mur_Artal_2015}. Another approach to achieve acceptable accuracy and robustness against various conditions is to use a range of complementary sensors.
	
	A Dynamic Vision Sensor (DVS) such as \cite{dvs_2008} is a visual sensor in which each pixel detects the log-intensity change at the pixel location and outputs an \emph{event} independent of the neighboring pixels. Event cameras are ideal for robotic applications because they allow low latency, high event generation rates, high dynamic range, and low power consumption.
	
	After surveying the related work in Section \ref{sec:related:work}, a short review of event data and event generation mechanism is presented in Section \ref{sec:event:data}. We then explain different components of the proposed algorithm in Section \ref{sec:algorithm} in detail. Section \ref{sec:experiments} evaluates the performance of the proposed algorithm. Finally, we conclude our discussion in Section \ref{sec:conclusion}.
	
	\section{Related Work} \label{sec:related:work}
	
	Since we represent events as intensity images and then track detected features, here we review both classical feature-based SLAM and event-based algorithms. Our aim here is to consider only the most relevant work rather than being comprehensive and complete.
	
	PTAM \cite{ptam_2007} is a feature-based algorithm that separates concepts of localization and mapping and 
	parallelizes both tasks in concurrent threads. This technique not only improves the overall performance but also provides more flexibility; the estimated parameters from one module can be exploited in another without waiting for the process to finish.
	
	ORB-SLAM \cite{orb_slam_Mur_Artal_2015} gradually built on PTAM to address some issues. The last version of this algorithm at the time of this writing, ORB-SLAM3 \cite{orbslam3_2021}, supports a range of sensors including regular cameras and the inertial measurement unit (IMU) with different configurations (monocular, stereo, fisheye, distorted pinhole). This algorithm uses the estimated map to localize pose, supports multi-session mapping, can relocalize when tracking is lost, and reduce the accumulated error with the \emph{loop closing} method. ORB-SLAM performance is limited due to the specifications of regular cameras, despite its robustness and accuracy.
	
	VINS-Mono \cite{vins_mono_2018} and its enhanced version, VINS-Fusion \cite{vins_fusion_2019}, are other feature-based methods that fuse different sensors to estimate the pose and structure using a window-based optimization scheme. Unlike ORB-SLAM, which associates features with their descriptors, VINS-Fusion is a KLT-based pipeline. Although it outperforms ORB-SLAM in some situations \cite{orbslam3_2021}, it still suffers from the same restrictions.
	
	In \cite{Gallego_2022_eventCamera_survey}, Gallego et al. surveyed event-based vision and its applications. In the case of event-based odometry and SLAM, researchers have begun by simplifying assumptions for the camera motion or scene structure. For instance, Gallego et al. \cite{accurate_ang_vel_gallego} proposed an algorithm for angular velocity estimation in which only pure 3D rotations were considered. However, most practical applications require more degrees of freedom, and a broader motion model should be considered.
	
	Mueggler et al. \cite{continuous_time_TE_Mueggler2015} proposed a continuous framework for pose estimation based on event data only. The information of every individual event is involved in state estimation. As noted in \cite{ultimate_slam_vidal_2018}, each event contains little information, so it is better to process a group of events. Additionally, some implementations of the continuous algorithms demand estimating many parameters in a very narrow time window.
	
	In \cite{sim_mosaicing_kim_2014}, Kim et al. reconstructed intensity images using event data and the features were detected and tracked in these frames. Although the algorithm performs well under fast motion and poor lighting conditions, it needs GPUs for real-time performance.
	
	Rebecq et al. \cite{ev_vi_rebecq_2017} used IMU measurements and events to track 6-DOF camera motion in a window-based optimization framework. They reconstructed motion-compensated images from consecutive overlapped spatiotemporal event windows. Then the features were extracted and tracked using the KLT \cite{klt_tracker_1981} method. Similarly, Vidal et al. \cite{ultimate_slam_vidal_2018} further utilized intensity images. Unlike \cite{ev_vi_rebecq_2017}, event windows were synchronized with the timestamp of intensity images. This synchronization can obscure the advantages of asynchronous low-latency event data. Both schemes require IMU measurements to initialize motion parameters and reconstruct frames. They also manually set the event window size for each sequence, which restricts the flexibility of their algorithm.
	
	Similar to our algorithm, the event-based visual-inertial odometry pipeline (EVIO) in \cite{evio_2017_zhu} adaptively selects the best spatiotemporal event window length based on the events' optical flow. However, while the proposed method depends only on input events in its basic configuration and addresses a range of conditions, EVIO relies on IMU readings for state estimation and does not explicitly consider situations where, for example, the camera is motionless. Unlike most event-based state estimation algorithms that estimate the pose and map parameters concurrently in an optimization framework, the proposed algorithm uses the reconstructed map to localize the current pose and is more like the ORB-SLAM algorithm in this sense.
	
	Recently, several multi-modal event-based algorithms have been proposed for 6DoF motion estimation. For example, \cite{crossmodal_zuo_2024} describes a method that uses events and a pre-generated map to track the camera. The map is either available or reconstructed using other sensors, such as regular cameras. This approach restricts its online performance, and although a map generated using depth cameras can tolerate illumination conditions or dynamics to a certain degree, spatiotemporal calibration of various sensors is critical for optimal performance.
	
	Similar to our algorithm, \cite{mc_veo_2024} uses a completely event-based motion-compensated approach to reconstruct event frames but estimates pose by registering these frames and intensity images. Although they use only one formulation to maximize event frame contrast iteratively, we believe a simpler and faster model is applicable to capture motion changes in certain conditions. Hence, we propose a variety of models to generate motion-compensated images. Moreover, this approach still suffers from the synchronization problem associated with registering intensity images and event windows, while the proposed algorithm modifies event window sizes adaptively to optimize its performance during adverse conditions, like when the camera moves slowly.
	
	In this paper, we propose an event-based SLAM algorithm with the following features:
	\begin{itemize}
		\item A novel image reconstruction algorithm adaptively selects the event window size based on camera motion and scene structure and converts them into a motion-compensated image.
		\item A higher-level KLT-based localization and mapping module exploits map data derived from MC images to estimate the current pose of the camera.
		\item An event-inertial version of the proposed event-only pipeline is presented to show how additional sensors can improve the algorithm. 
		\item Distorted pinhole and Kannala Brandt \cite{kannala_brandt_cam_model} camera models are supported, the camera motion has six degrees of freedom (6-DOF), and there is no assumption about the type of scene.
		\item The proposed algorithm has robust performance in different conditions as long as the map estimate is reliable and there is relative motion between the camera and the scene. 
		\item The code of this project can be found at \cite{eorbslam_github_url}.
	\end{itemize}

	\section{Event Data} \label{sec:event:data}
	
	If there is a relative motion between the DVS and the scene, 
	the illumination $I$ at each pixel location changes. 
	When there is enough change in the log-intensity, $L(t_k) = \log (I(t_k))$ at current time $t_k$, 
	relative to a reference time $t_r$, it outputs an event at each pixel coordinate $\pmb{x}_k = (x_k, y_k)$. 
	Formally, if the change in $L$ is greater than a threshold, $C$, 
	\begin{equation}
		| \Delta L(t_k, t_r) | = | L(t_k) - L(t_r) | > C
	\end{equation}
	there is an event
	$$ \pmb{e}_k: \lbrace t_k, x_k, y_k, p_k \rbrace $$
	where $p_k$ shows the sign of change, i.e., if $L(t_k)$ is bigger than $L(t_r)$, 
	$p_k$ is positive, and it is negative otherwise.
	
\begin{figure}[htbp]
		\centering
		\begin{subfigure}{.25\textwidth}
			\centering
			\includegraphics[width=.85\textwidth]{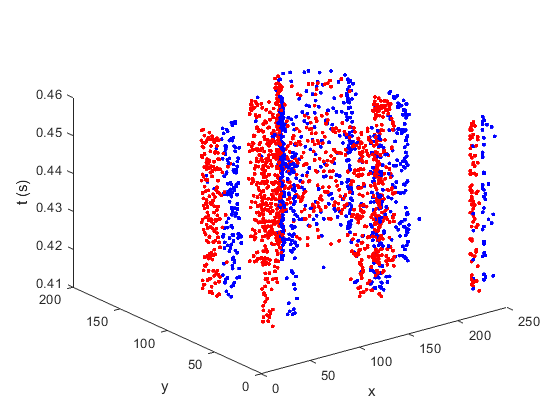}
			\caption{}
			\label{fig:eventsAndFrames:evs3d}
		\end{subfigure}%
		\begin{subfigure}{.25\textwidth}
			\centering
			\includegraphics[width=.85\textwidth]{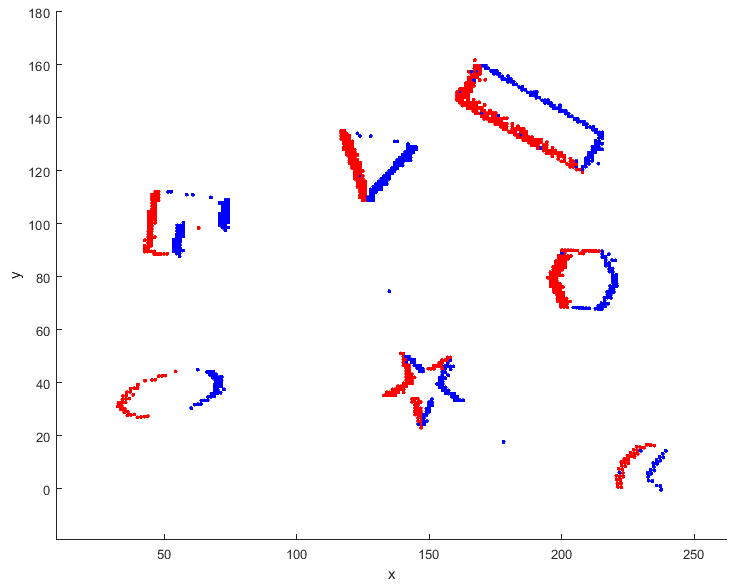}
			\caption{}
			\label{fig:eventsAndFrames:evs2d}
		\end{subfigure}
		\caption{%
			Representation of an event window from shapes\_6dof \cite{ethz_pub_ds_Mueggler_2017}. 
			(\subref{fig:eventsAndFrames:evs3d}) A window of 2000 events starting from a specific timestamp. Blue and red points show events with positive and negative polarity, respectively.
			(\subref{fig:eventsAndFrames:evs2d}) 2D view of (\subref{fig:eventsAndFrames:evs3d}) in the image ($x$-$y$) plane.
		}
		\label{fig:eventsAndFrames}
	\end{figure}

	Fig. \ref{fig:eventsAndFrames} shows a slice of all events in the sequence 
	shapes\_6dof from the Public Event Dataset \cite{ethz_pub_ds_Mueggler_2017}. 
	Each 3D point in the spatiotemporal space in Fig. \ref{fig:eventsAndFrames}(\subref{fig:eventsAndFrames:evs3d}) designates an event, color-coded based on event polarities. Each slice of this space represents an event window. 
	
	Since event generation depends on intensity changes at each pixel, there is no spatial correlation between adjacent pixels. Each pixel generates events asynchronously as soon as a change in its intensity value is detected.
	On the other hand, uniform areas in the image cannot contribute to event generation.
	This redundancy reduction allows lower latencies and faster event generation rates. 
	Events compactly summarize information regarding the high-contrast areas of the scene.
	
	Note how the event generation rate depends on the composition of the scene and the relative motion of the camera in normal illumination conditions. For a fixed threshold, $C$, if the camera traverses slowly through the environment, the event generation rate is low, and event data is noisy. On the other hand, in highly textured scenes, the event generation rate could be high at regular speeds. For specific camera settings, if the camera moves fast in a textured environment, the event generation rate eventually reaches the maximum allowable output bandwidth of the event camera. We do not explicitly consider the degrading effects of extreme illumination changes in this study.
	
	We can benefit from available image-based SLAM algorithms by representing event data as a 2D image.
	To produce the event histogram $I_r$ for a specific event window $W_r$ at reference time $t_r$, 
	all events in each pixel location are added
	\begin{equation}
		\label{eq:mcImage}
		I_r(\pmb{x}) = \sum _{\pmb{e}_k \in W_r} p_k \delta (\pmb{x} - \pmb{x}_k^\prime),
	\end{equation}
	where $\delta (.)$ is the continuous Dirac's delta kernel. 
	It is common to replace the continuous delta operator with a discrete-time sampled Gaussian kernel with an arbitrary standard deviation $\sigma _I$.
	The parameter $p_k$ can be changed based on event polarity or set to constant 1 for all events in the window.
	
	Depending on the camera speed and the length of the event window, the event histograms may suffer from motion distortion. In this case, event coordinates can be corrected before reconstructing these images. The resulting image is called a Motion-Compensated Image or MCI for short.
	
	For a carefully selected event window, we can mitigate the motion distortion using a two-dimensional similarity or rigid body transformation [Sim(2) or SE(2)]. 
	Assuming a constant rotational speed $\omega$ and translational speed $\pmb{v} = (v_x, v_y)$ for each event $\pmb{e}_k$ in the window, the warp
	\begin{equation}
		\label{eq:ev2dWarp}
		\begin{bmatrix}
			x^\prime \\
			y^\prime
		\end{bmatrix} = s
		\begin{bmatrix}
			\cos \theta _k & -\sin \theta _k \\
			\sin \theta _k & \cos \theta _k
		\end{bmatrix}
		\begin{bmatrix}
			x \\
			y
		\end{bmatrix} +
		\begin{bmatrix}
			t_{k,x} \\
			t_{k,y}
		\end{bmatrix}
	\end{equation}
	maps the event location $\pmb{x}$ at $t_k$ to $\pmb{x} ^\prime$ at time $t_r$ where $\theta _k = \omega \Delta t$,
	$\pmb{t} _k = \pmb{v} \Delta t$, $s$ is the arbitrary scale, and $\Delta t = |t_k - t_r|$. 
	Event coordinates should be normalized based on the camera intrinsics beforehand. Also, note that $\Delta t$ can be varied, but this does not cause any issues as long as the constant speed assumption holds.
	
	When the relative transform $T_{t_r,t_k}$ and the depth $Z(\pmb{x}_k)$ of each event in the window are known, a SE(3) mapping can be used to warp events: 
	\begin{equation}
		\label{eq:ev3dWarp}
		\pmb{x}_k^\prime = \pi _0(T_{t_r,t_k}\lbrack Z(\pmb{x}_k)\pi _0^{-1}(\pmb{x}_k)\rbrack).
	\end{equation}
	Here $\pi _0(.)$ projects the 3D point to the image plane viewed from the current pose, and $\pi _0^{-1}(.)$ is its inverse operation.
	In this case, we first retrieve the 3D feature point associated with each event, seen from the current camera position, and map it to a reference location.
	
	If we extract feature points from the reference reconstructed frame and track them across the current image, it is possible to estimate the parameters of (3) and (4) for all events between these frames. For this, we minimize the reprojection error between each warped reference feature location $\pmb{x}_i$ and the corresponding match $\pmb{x}_j$ in the current frame using
	\begin{equation}
		\label{eq:argminEq}
		\pmb{\theta}^* = \underset{\pmb{\theta}^*}{\text{argmin.}} \sum_{i,j \in J}^{} \pmb{e}_{ij}^T(\pmb{x}) \pmb{\Omega}_{ij} \pmb{e}_{ij}(\pmb{x}),
	\end{equation}
	where $\pmb{\theta}^*$ is the transformation parameters, $\pmb{e}_{ij} = \pmb{x}_i - \pmb{x}_j$, $\pmb{\Omega}_{ij}$ is the information matrix, and $J$ is the set containing all matched feature pairs between the two frames.
	
	To use the estimated parameters $\pmb{\theta}^*$ to warp events in the current window, we first convert the relative transform to speed, $\pmb{\psi}$, using 
	\begin{equation*}
		\pmb{\psi} = \frac{\text{Log}(\pmb{\theta}^*)}{\Delta T},
	\end{equation*} 
	where $\Delta T$ is the length of the window in units of time and $\text{Log}(.)$ is the inverse algebra that maps an element of SE(2) or SE(3) to a member of the corresponding tangent space. On the premise that the camera moves slow enough to assume constant speed for all events, the unknown transform between each event location and the reference timestamp, $T_{t_r,t_k}$, is 
	\begin{equation*}
		T_{t_r,t_k} = \text{Exp}(\pmb{\psi} \Delta t),
	\end{equation*} 
	where $\Delta t = | t_k - t_r |$ and $\text{Exp}(.)$ is the exponential map for the corresponding Lie group.
	
	We also distinguish the notion of an image and a frame. While an image is simply a two-dimensional array of numbers, a frame consists of other information, including the timestamp, camera pose, and extracted key points.
	
	Interested readers can refer to \cite{Gallego_2022_eventCamera_survey} for an in-depth review of the event generation mechanism and event representation.

	\section{Algorithm} \label{sec:algorithm}
	
	\begin{figure*}
		\centering
		\includegraphics[trim=20 110 40 15,clip,width=.85\textwidth]{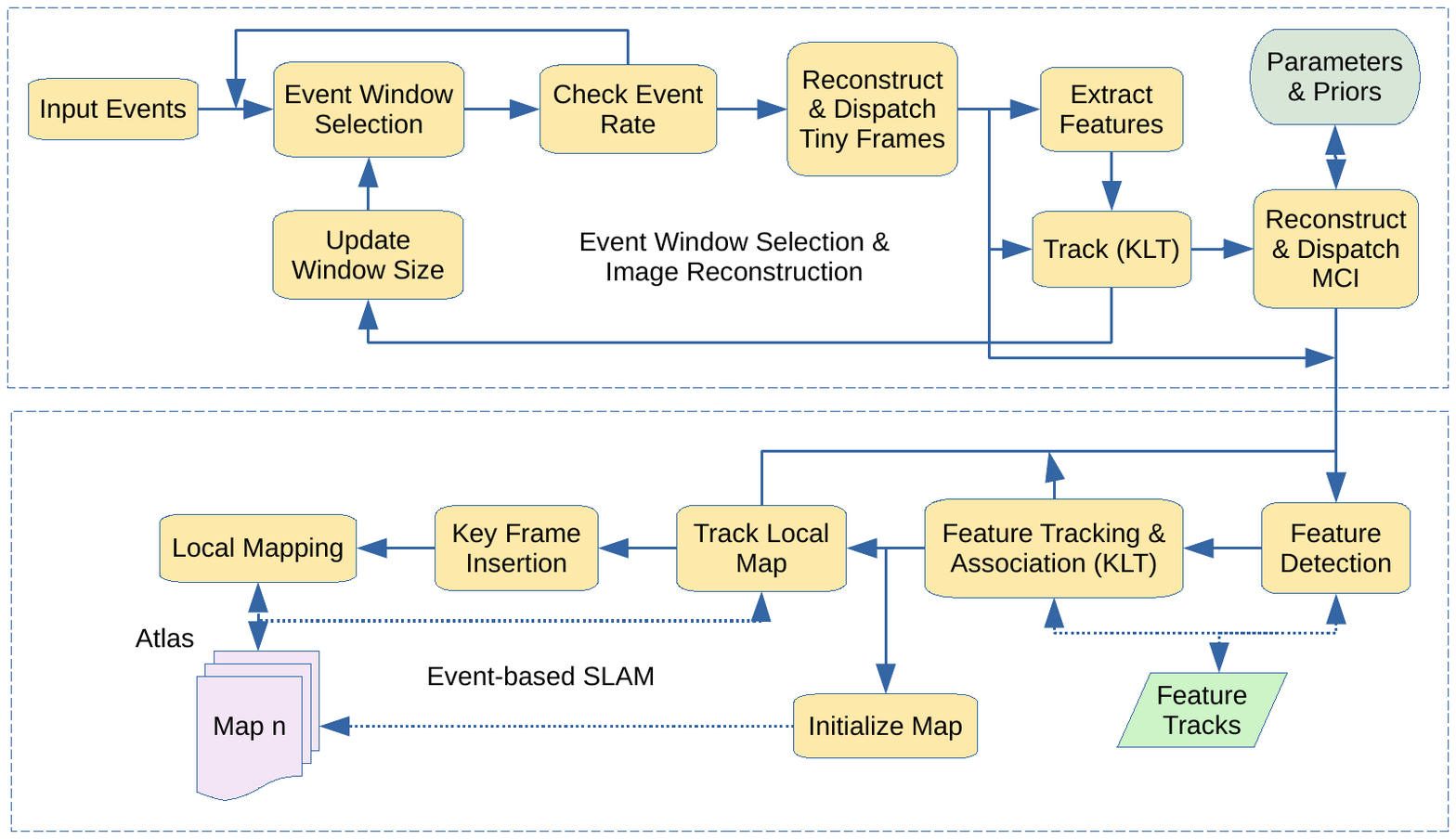}
		\caption{An overview of the proposed event-based SLAM pipeline.}
		\label{fig:aekltBlockDiagram}
	\end{figure*}
	
	Fig. \ref{fig:aekltBlockDiagram} illustrates an overview of the proposed event-based SLAM algorithm. Our pipeline consists of two main components, which run concurrently in separate threads. The first part selects an appropriate spatiotemporal window of input events and reconstructs an image for each event window. The second part of the proposed algorithm extracts features from each input image and tracks them using the KLT method across the following frames. The 6-DOF pose of the camera is then estimated, and the local scene is reconstructed. Sections \ref{subsec:l1ImageReconst} and \ref{subsec:l2EvTracking} discuss each component in more detail.
	
	\subsection{Event Window Selection and Image Reconstruction} \label{subsec:l1ImageReconst}
	
	\begin{algorithm}[ht]
		\caption{Event Window Selection and Image Reconstruction}
		\label{alg:evImgGen}
		\begin{algorithmic}[1]
			\REQUIRE Event stream $\{\,\pmb{e}_k  (t_k, x_k, y_k, p_k)\;\forall \; k = 0, 1, \dots\,\} $ and parameters ($N_e$, $th_{N_f}$, $th_e$, $th_{fd}$)
			\ENSURE Reconstructed images (tiny images and MCIs) and motion priors
			
			\FOR{each window of $N_e$ events}
			\STATE calculate $r$ from \eqref{eq:evGenRate}
			\IF{$r < th_e$}
			\STATE reconstruct tiny image \eqref{eq:mcImage}
			\STATE extract FAST features 
			\IF{enough features detected}
			\IF{reference frame is not initialized}
			\STATE Initialize reference frame
			\ELSE
			\STATE track reference features in current frame (KLT optical flow)
			\STATE test two-view reconstruction with RANSAC outlier rejection
			\STATE calculate median feature displacement \eqref{eq:medFtDisp}
			\IF{$M_{fd} > th_{fd}$}
			\STATE \emph{MCI\_Gen}: reconstruct and dispatch MCI \\(Algorithm \ref{alg:mciGen})
			\STATE update $N_e$ \eqref{eq:neUpdate}
			\STATE reset the reference frame
			\ENDIF
			\ENDIF
			\ENDIF
			\ELSIF{$N_{\text{tiny frames}} >= th_{N_f}$}
			\STATE \textbf{goto} \emph{MCI\_Gen}
			\ENDIF
			\ENDFOR
		\end{algorithmic}
	\end{algorithm}
	
\begin{figure}[htbp]
		\centering
		\includegraphics[width=.45\textwidth]{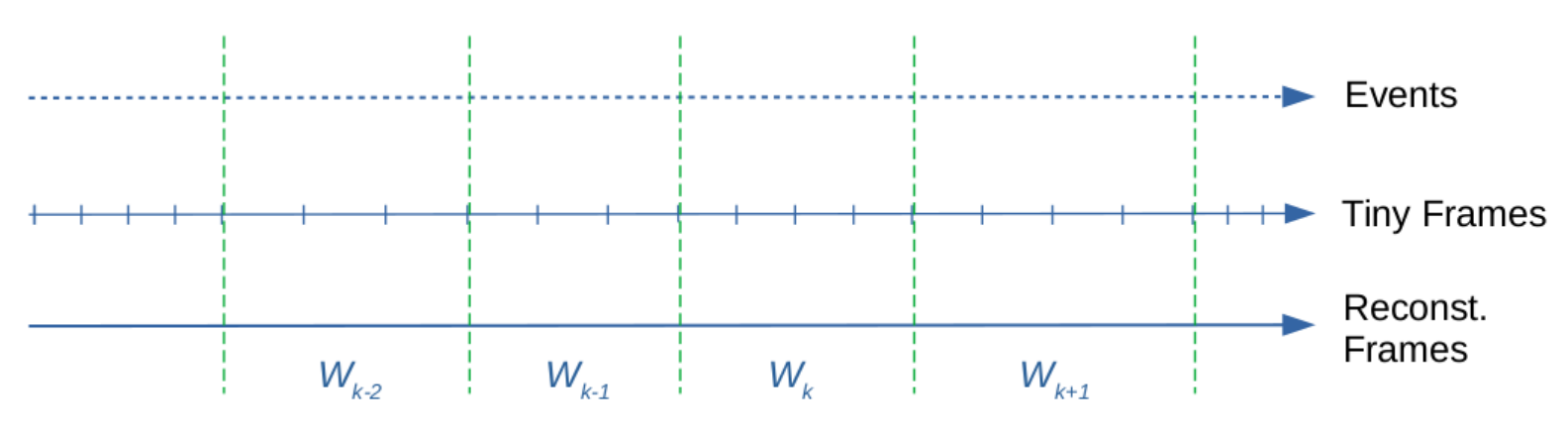}
		\caption{An illustration of the input event stream, tiny event windows, and non-overlapped reconstruction windows relative to each other.}
		\label{fig:eventWindowsRel}
	\end{figure}

	The basic idea of the algorithm is to process events in small chunks and generate an MCI whenever there are enough events (Algorithm \ref{alg:evImgGen}).
	The algorithm uses two sets of event windows to select the appropriate window size and reconstruct the MCI. It continually tracks and accumulates \emph{tiny windows} to determine whether there are enough events. Then, it reconstructs the MCI using the \emph{reconstruction window} consisting of all previously collected events.
	Fig. \ref{fig:eventWindowsRel} illustrates the relationship between the event stream, tiny windows, and reconstruction windows. Note that a reconstruction window contains a fixed number of tiny windows, and the window sizes are updated according to the conditions described below.
	
	This algorithm first selects a window of input events.
	Initially, the length of the window, $N_e$, is selected arbitrarily and can be adjusted later. At this stage, consecutive event windows have no overlap. 
	The proper event window size depends on the event generation rate. As discussed in Section \ref{sec:event:data}, 
	the maximum bandwidth of the event camera limits the event generation rate. For example, in an event camera with a bandwidth of 1 million events per second and a resolution of $240\times180$ pixels, we can expect up to around 23 events per second per pixel. Thus the minimum event generation rate can be chosen based on the bandwidth of the event camera, e.g., one event per pixel per second. We select this threshold solely based on the event camera specifications agnostic to the camera motion and scene.
	
	Based on these observations, the event generation rate is calculated by
	\begin{equation}
		\label{eq:evGenRate}
		r = \frac{N_e}{\Delta t W H}
	\end{equation}
	where $\Delta t = t_{e_{N_e}} - t_{e_1}$ is the difference between the first event timestamp, $t_{e_1}$, and the last timestamp, $t_{e_{N_e}}$ in the current window, and $W$ and $H$ are the width and height of the event image in pixels, respectively.
	
	If the camera does not move fast enough and the event generation rate is low then the event data will be noisy. The algorithm rejects most tiny windows with an event generation rate less than a threshold, $th_e$, and restarts from the initial state. The only exception is when there are at least $th_{N_f}$ frames. In this case, the algorithm restarts after it generates and dispatches the MCI.

	If the event generation rate of the current tiny window is acceptable, a tiny frame is reconstructed based on \eqref{eq:mcImage} without motion compensation. The following sections review the main components of the image reconstruction algorithm.
	
	\subsubsection{KLT Optical Flow Initialization and Tracking}
	
	In each iteration, the algorithm detects FAST features \cite{fast_features_2006} in the first tiny frame. Since we use image features to determine the displacement in event locations, fewer features are sufficient. 
	
	The algorithm then tracks the reference features in the subsequent frames. We can use the feature matches obtained from the feature tracks for the two-view reconstruction.
	The map initialization algorithm in \cite{orb_slam_Mur_Artal_2015} is used to estimate the relative pose between the current and the reference frames and the 3D map points corresponding to the feature matches.
	Due to the small baseline between the consecutive tiny frames, most scene reconstruction algorithms cannot yield reliable results.
	So, we relieve some of the stringent conditions in \cite{orb_slam_Mur_Artal_2015}, i.e., the best model does not have to stand out by a large margin.
	Because most implementations use the RANSAC method internally, feature matches are still enhanced despite potential failures in scene reconstruction.
	If a successful reconstruction of the 3D structure is possible, this information is stored and used in the MCI reconstruction step. 

	We define the median feature displacement metric as the median distance of the tracked image features in pixels
	\begin{equation}
		\label{eq:medFtDisp}
		M_{fd} = \text{median}(\{||\pmb{x}_{ij} - \pmb{x}_{rj} ||_2 \; \forall j = 0, 1, \dots\})
	\end{equation}
	where $\pmb{x}_{rj}$ is the location of the $j$th feature in the reference frame, and $\pmb{x}_{ij}$ is the corresponding match in the current frame. Whenever the median feature displacement in the current tiny frame is higher than a threshold, $th_{fd}$ (e.g., five pixels), enough events have been accumulated, and the upcoming MCI reconstruction is commenced.
	
	Note that depending on the type of the scene, the number of detected features varies, but the median feature displacement is not affected. We select the feature displacement threshold based on how much distortion we can tolerate in event locations. This threshold is constant during the operation of the algorithm for all scenes.
	
	Before the algorithm begins the next iteration, it updates the size of the tiny windows, $N_e$. If $N_f$ is the number of processed frames and $N_x$ is the number of expected frames per iteration, the new $N_e^*$ is 
	\begin{equation}
		\label{eq:neUpdate}
		N_e^* = \big\lfloor \frac{N_f N_e}{N_x} \big\rfloor,
	\end{equation}
	where $\lfloor x \rfloor$ is the greatest integer less than or equal $x$.
	
	In this case, $N_e$ is adjusted according to the camera speed automatically. If the camera speed increases, fewer events satisfy the feature displacement condition, and $N_e$ decreases. On the other hand, if the camera moves slowly, we need more events for a defined feature displacement, and $N_e$ increases. Since we did not assume any scene structure,  $N_e$ is independent of the type of the scene.
	
	\subsubsection{Motion Compensated Images}
	
	We use different warps to rectify motion distortion in event coordinates and use \eqref{eq:mcImage} to reconstruct the MCI. We consider the last event timestamp as the reference and project preceding events in the forward direction.
	
	If there is a reliable 3D reconstruction, we exploit this information in \eqref{eq:ev3dWarp} to warp the event locations. As in \cite{ultimate_slam_vidal_2018}, the median depth of 3D map points is used for all events in the window. Prior to this operation, motion and map parameters can be enhanced using the Bundle Adjustment (BA) method.
	
	On the other hand, if there is no initial scene and motion estimation, but we have tracked the features across several tiny frames, we can still undistort events using a 2D optimization scheme. In this case, we first fit a Sim(2) or SE(2) motion model to the feature tracks and then reproject events to the reference frame according to \eqref{eq:ev2dWarp}. We also store this model as initialization parameters for the subsequent iterations.
	
	\begin{algorithm}[t]
		\caption{MCI Reconstruction}
		\label{alg:mciGen}
		\begin{algorithmic}[1]
			\REQUIRE Reconstruction event window and motion priors
			\ENSURE MCIs and refined motion priors
			
			\STATE Test several hypothesises concurrently:
			
			\IF{3D reconstruction available}
			\STATEx \; \textbf{H1}: Enhance motion and map parameters using Bundle Adjustment (BA) [optional]
			\STATEx \;\; Reconstruct using \eqref{eq:ev3dWarp} and median depth for all events
			\ENDIF
			\STATEx \; \textbf{H2}: Fit a Sim(2) or SE(2) motion model to feature tracks (use previous estimations as initialization point)
			\STATEx \;\; Reconstruct using \eqref{eq:ev2dWarp}
			\STATEx \; \textbf{H3}: Estimate average depth using \eqref{eq:argminEq} and pose priors (e.g. from IMU integration or last pose estimate)
			\STATEx \;\; Reconstruct using \eqref{eq:ev3dWarp}
			\STATEx \; \textbf{H4}: Reconstruct the event histogram with no motion compensation \eqref{eq:mcImage}
			\STATE Select and dispatch the image with the highest average local STD
			
		\end{algorithmic}
	\end{algorithm}
	
	In some situations, only the relative pose estimate is known. When the camera motion is adequately slow, we can either integrate the IMU measurements or use the last pose estimate to infer the current camera location. In either case, we can estimate the average depth of the scene using \eqref{eq:argminEq} and warp event locations based on \eqref{eq:ev3dWarp}.
	
	We run each reconstruction method concurrently in parallel threads. Besides the MCIs, we reconstruct the event histogram using \eqref{eq:mcImage} with no motion compensation. Finally, we score each image based on its contrast or \emph{sharpness} \cite{mcImage_allego_2018}. To measure image sharpness, we first divide each image into non-overlapped patches and compute the local standard deviation (STD) for each patch. We select and dispatch the image with the highest average local STD. We choose the local STD because it yields more stable results than the global STD.
	
	The KLT algorithm has a poor performance when there is a large baseline between consecutive frames \cite{klt_tracker_1981}. Although there is not much motion between tiny frames, the distance between the reconstructed frames can be much higher. One way to overcome this issue is to send tiny frames in addition to MCIs to the event-based tracking algorithm (Section \ref{subsec:l2EvTracking}). 
	
	Another approach to ensure smooth transition between consecutive MCIs is to overlap the reconstruction windows. 
	We prefer this method to the non-overlapped scheme because it leads to more robust feature tracking at the second level.
	
	\subsection{Event-based Localization and Mapping} \label{subsec:l2EvTracking}
	
	With the MCI from the previous step, the event-based localization module follows an image-based KLT optical flow scheme to estimate the 6-DOF pose of the camera and reconstruct the scene. The basic steps of this algorithm are almost similar to the reconstruction algorithm discussed in Section \ref{subsec:l1ImageReconst}.
	
	\subsubsection{Initialization}
	
	If there are not enough feature tracks, the algorithm detects new FAST features in the input MCI. It then creates and merges new feature tracks to manage feature locations across successive images. We use a bucketing grid scheme to ensure a uniform distribution of detected features across images.
	
	The last features are tracked in each subsequent image using the KLT method. If the map is not initialized, we use the two-view reconstruction algorithm to reconstruct the scene and recover the camera's relative motion between the current and reference frames. As in Section \ref{subsec:l1ImageReconst}, we use the map initialization algorithm in \cite{orb_slam_Mur_Artal_2015} for this step. After a successful reconstruction, we initialize the map and perform a global bundle adjustment optimization to enhance the estimation.
	
	\subsubsection{Tracking The Local Map}
	
	We can use feature tracks to identify feature and map point associations across different frames. Therefore, the geometrical distance between the projected map points in the current frame and the corresponding key point observations is minimized to estimate the current pose of the camera.
	
	Due to the camera motion, we might lose track of some features. In this case, the algorithm fails to track features that are not visible in the subsequent frames.

	\subsubsection{Key Frame Insertion and Local Mapping}
	
	The algorithm inserts a new keyframe whenever there is enough baseline between the current tracked frame and the last keyframe. We use several heuristics to decide when to spawn a new keyframe. Whenever the number of tracked map points falls below a threshold or the median pixel displacement between key points in the current frame and the preceding keyframe is greater than a certain limit, a new keyframe is created.
	
	The local mapping module is very similar to the method used in ORB-SLAM. While our algorithm uses the KLT method to track and associate features, ORB-SLAM detects new matches with descriptors. 
	
	In local mapping, we first cull outlier map points based on their rate of observations. Additional map points are then triangulated for the new feature matches between the current and previous keyframes. Finally, a local bundle adjustment optimization enhances the map point estimates.

	\subsection{IMU Measurements}
	
	In this section, the necessary changes to incorporate IMU measurements are introduced. The preintegration theory and the IMU initialization module of \cite{imu_preintegration_orster_2017} and \cite{orb_slam_Mur_Artal_2015} are used to accomplish this goal. These changes are as follows:
	
	\begin{itemize}
		\item IMU measurements are integrated between the tiny frames, MCIs, and the keyframes.
		\item In the MCI reconstruction module, using the initial pose from the IMU integrations and feature matches, we first find the average depth of the scene in the 3D warp \eqref{eq:ev3dWarp} through the optimization framework of \eqref{eq:argminEq}. We also consider IMU biases and the gravity direction as the state variables in this optimization. The initial gravity direction is evaluated using the first accelerometer reading for the current iteration. These parameters are stored for the subsequent iterations, and the MCI is reconstructed. 
		\item We use an inertial bundle adjustment optimization framework in the MCI generation step with a successfully reconstructed scene. In addition to the IMU biases and the gravity direction, we also consider the relative scale. 
		\item We adapt the inertial local bundle adjustment and the mechanism of IMU initialization discussed in \cite{orb_slam_Mur_Artal_2015} in the local mapping algorithm. In summary, the IMU biases, the direction of the gravity, and the relative scale are estimated using an adequate number of keyframes. Such optimization and scale refinement are repeated for several predefined periods. 
	\end{itemize}

	\section{Experiments} \label{sec:experiments}
	
	To evaluate the performance of the proposed pipeline, we use three publicly available event datasets,
	the Public Event \cite{ethz_pub_ds_Mueggler_2017}, the Multi-vehicle Stereo Event \cite{mvsec_ds_Zhu_2018}, and Stereo DAVIS Dataset \cite{stereo_davis_dataset_2018}. The former includes short-duration sequences recorded in different scenes and challenging conditions by a monocular 240$\times$180 pixel DAVIS device. An embedded IMU measures acceleration and angular velocity along three axes of freedom. Most sequences include events, intensity images, IMU measurements, and the ground truth. We only consider the ones for which both the ground truth and IMU measurements are available; hence we exclude sequences from the ``depth'' group.
	
	Sequences of \cite{mvsec_ds_Zhu_2018} are recorded with two similar DAVIS 346$\times$260 pixel cameras attached in the stereo configuration. It contains events and intensity images for both cameras, IMU measurements, the ground truth trajectory, and depth maps for each frame. We only use the left camera and several representative outdoor sequences, noting that the proposed pipeline is not designed to tackle the stereo case or track longer outdoor distances. Finally, \cite{stereo_davis_dataset_2018} uses two DAVIS cameras similar to the first dataset in stereo configuration. Although IMU data are provided in addition to the events, images, and ground truth, IMU-camera calibration parameters are not readily available.
	
	We use several objective criteria to verify the results of the proposed algorithm quantitatively. Since our algorithm generally results in an atlas of multiple disconnected pose graphs, we extend the Relative Pose Error (RPE) in \cite{tum_rgbd_sturm12iros} to average the normalized relative errors as 
	\begin{equation}
		\overline{RPE}(\pmb{\theta}) = \frac{1}{|\mathcal{A}|} \sum_{\mathcal{G} \in \mathcal{A}} \frac{1}{D |\mathcal{G}|} \sum_{i,j \in \mathcal{G}} \pmb{\theta}(\pmb{s}(\Delta T^*_{i,j}) \ominus \Delta T_{i,j})
	\end{equation}
	where $\mathcal{A}$ is the set of all pose graphs, $\mathcal{G}$, 
	$\Delta T_{i,j} = T_i \ominus T_j$ is the relative transform between pairs of SE(3) poses in $\mathcal{G}$, $T_i$ and $T_j$, 
	$\ominus$ is the inverse of SE(3) Lie algebra, $D$ is the total traversed distance for the current graph, $\pmb{s}()$ is the scaling operation for monocular event-only tracking, $\pmb{\theta}()$ returns either rotational or translational component of error, and ${}^*$ indicates the estimate. We normalize RPE by the total traversed distance to reflect the effects of variable length pieces. For the monocular event-only configuration, we calculate the unknown scale by comparing the associated estimate and ground truth pairs and scale the estimate before computing the mean error. 
	
	Additionally, we consider measures to assess the stability of the pipeline. Total traversed distance or time is the sum of all distances or delta times between consecutive frames in each pose graph, and is given by
	
	\begin{equation}
		\label{eq:stability}
		\tau(\pmb{\theta}) = \sum_{\mathcal{G} \in \mathcal{A}} \sum^{N_{\mathcal{G}} - 1}_{\substack{i = 0 \\ i \in \mathcal{G}\{N_{\mathcal{G}}\}}} |\pmb{\theta}_{i+1} - \pmb{\theta}_{i} |
	\end{equation}
	where $\tau(.)$ is the stability metric, $\mathcal{G}$ is a pose graph with $N_{\mathcal{G}}$ elements, and $\pmb{\theta}$ can either be the timestamp or the 3D position of the $i$th pose estimate.
	
	Before the algorithm processes input events, lens distortions in event pixel locations are rectified using available calibration parameters for each sequence. We choose $N_{x} = 3$, and the initial value of $N_{e}$ is 2000 events for sequences of \cite{ethz_pub_ds_Mueggler_2017} and \cite{stereo_davis_dataset_2018} and 6000 for \cite{mvsec_ds_Zhu_2018}. 
	Since the event frames in \cite{mvsec_ds_Zhu_2018} have a higher resolution, a higher value for $N_{e}$ helps the convergence speed, though starting from 2000 events should eventually converge to the optimal value.
	Motion-compensated windows have 50\% overlap, and
	Event frames are reconstructed using a Gaussian kernel with $\sigma _I = 1$ in both tracking levels (tiny frames and MCIs). The FAST feature detector threshold is set to a small amount (around zero) because this setting yields the most features and allows us to filter them by their response. We set the KLT tracker with two pyramid levels, a block size of 23$\times$23 pixels, and a maximum bidirectional error of one. These settings are chosen arbitrarily based on our experiments%
	\footnote{Video representation can be found on: https://youtu.be/rKIBkMns5No}%
	.
	
	\begin{table*}[t]
		\caption{Performance evaluation of the proposed algorithm for two event-based configurations compared against the ground truth for some sequences of \cite{ethz_pub_ds_Mueggler_2017} and \cite{mvsec_ds_Zhu_2018}.}
		\ra{1.3}
		\begin{center}
			\begin{tabular}{@{}llccccccc@{}}
				\hline
				\hline
				& & \multicolumn{3}{c}{\textbf{E-Only}} & & \multicolumn{3}{c}{\textbf{E-I-C1}} \\
				\cmidrule{3-5} \cmidrule{7-9}
				& & $\overline{\text{\textbf{\textit{RPE}}}}$ & $\overline{\text{\textbf{\textit{RPE}}}}$ & & & $\overline{\text{\textbf{\textit{RPE}}}}$ & $\overline{\text{\textbf{\textit{RPE}}}}$ & \\
				\textbf{\textit{Dataset}} & \textbf{\textit{Sequence}} & \textbf{\textit{Position}} & \textbf{\textit{Rotation}} & \textbf{\textit{Stability}} & & \textbf{\textit{Position}} & \textbf{\textit{Rotation}} & \textbf{\textit{Stability}} \\
				& & \textbf{\textit{(-)}} & \textbf{\textit{(deg/m)}} & \textbf{\textit{($\times 10^3 \text{m.s}$)}} & & \textbf{\textit{(-)}} & \textbf{\textit{(deg/m)}} & \textbf{\textit{($\times 10^3 \text{m.s}$)}}\\ 
				\hline
				& shapes\_6dof & 0.048 & 0.209 & 2.31 & & 0.050 & 0.114 & 2.50 \\ 
				& shapes\_translation & 0.084 & 0.186 & 2.29 & & 0.092 & 0.169 & 2.62 \\ 
				Public & poster\_6dof & 0.017 & 0.062 & 3.18 & & 0.018 & 0.057 & 3.14 \\ 
				Event & poster\_translation & 0.005 & 0.006 & 2.29 & & 0.005 & 0.006 & 2.33 \\ 
				\cite{ethz_pub_ds_Mueggler_2017} & hdr\_poster & 0.011 & 0.025 & 2.67 & & 0.002 & 0.007 & 2.73 \\ 
				& boxes\_6dof & 0.002 & 0.004 & 3.91 & & 0.006 & 0.008 & 3.95 \\ 
				& boxes\_translation & 0.004 & 0.004 & 3.40 & & 0.004 & 0.004 & 3.46 \\ 
				& hdr\_boxes & 0.004 & 0.007 & 3.08 & & 0.004 & 0.006 & 3.07 \\ 
				& dynamic\_6dof & 0.019 & 0.041 & 2.09 & & 0.032 & 0.117 & 2.11 \\ 
				& dynamic\_translation & 0.012 & 0.018 & 1.24 & & 0.034 & 0.032 & 1.38 \\ 
				\hline
				& indoor\_flying1 & 0.044 & 0.041 & 1.12 & & 0.080 & 0.046 & 1.34 \\ 
				Multi- & indoor\_flying2 & 0.041 & 0.032 & 1.86 & & 0.029 & 0.019 & 2.11 \\ 
				Vehicle & indoor\_flying3 & 0.018 & 0.006 & 3.23 & & 0.029 & 0.008 & 3.63 \\ 
				Event & indoor\_flying4 & 0.038 & 0.013 & 0.11 & & 0.038 & 0.016 & 0.12 \\ 
				Stereo & outdoor\_day1 & 0.889 & 0.033 & 3.14 & & 1.181 & 0.041 & 6.38 \\ 
				\cite{mvsec_ds_Zhu_2018} & outdoor\_night1 & 0.678 & 0.003 & 3.29 & & 1.131 & 0.008 & 3.94 \\ 
				& outdoor\_night3 & 1.208 & 0.009 & 5.04 & & 1.406 & 0.017 & 8.95 \\ 
				\hline
				\hline
			\end{tabular}
			\label{tab:ev:configs:results}
		\end{center}
	\end{table*}
	
	Table \ref{tab:ev:configs:results} summarizes the evaluation results for event-only (E-Only) and event-inertial (E-I-C1) sensor configurations. The stability column of this table is the multiplication of the total traversed distance in meters and time in seconds ($\tau(t).\tau(\pmb{p})$). Although event-only tracking shows similar or superior results in most cases, the inertial method is more stable. Since more challenging periods can be tracked in the inertial case, the errors grow accordingly. Furthermore, because we do not scale inertial pose graphs, the reported results also include scaling errors.
	
	The inertial configuration discussed in Table \ref{tab:ev:configs:results} (E-I-C1) still assumes that the tracking can be lost due to severe conditions, so it tries to reinitialize the map as soon as possible. To further assess the limitations of the proposed algorithm and compare it against the inertial state estimation in \cite{ultimate_slam_vidal_2018}, we consider two other configurations, E-I-C2 and E-I-C3. We configure our pipeline to enforce continuous tracking without spawning and initializing new maps in unfavorable conditions. In both cases, we disallow the reinitialization of the map and continue to estimate 3D points using inertial readings and feature matches between the nearby frames. In the last configuration (E-I-C3), we also fix the tiny event window size to a predefined value and restrict the withdrawal of noisy frames based on the event generation rate. Table \ref{tab:ev:fixed:tf:sizes} summarizes our preferences for the length of fixed tiny windows for each group of sequences in \cite{ethz_pub_ds_Mueggler_2017}. We select these values based on our experience of E-I-C1 with variable window sizes.
	
	\begin{table}
		\caption{Fixed tiny event window size specification for each group of sequences in \cite{ethz_pub_ds_Mueggler_2017}.}
		\ra{1.3}
		\begin{center}
			\begin{tabular}{@{}ll@{}}
				\hline
				Sequence & Number of Events \\
				\hline 
				shapes & 2000 \\
				dynamic & 8000 \\
				poster & 10000 \\
				boxes & 12000 \\
				\hline 
			\end{tabular}
			\label{tab:ev:fixed:tf:sizes}
		\end{center}
	\end{table}
	
	\begin{table*}
		\caption{Comparison of the performance of different inertial configurations of the proposed algorithm for sequences of \cite{ethz_pub_ds_Mueggler_2017}.}
		\ra{1.0}
		\begin{center}
			\begin{tabular}{@{}lcccccccc@{}}
				\hline
				\hline
				& \multicolumn{2}{c}{\textbf{E-I-C1}} & & \multicolumn{2}{c}{\textbf{E-I-C2}} & & \multicolumn{2}{c}{\textbf{E-I-C3}} \\
				\cmidrule{2-3} \cmidrule{5-6} \cmidrule{8-9}
				& $\overline{\text{\textbf{\textit{RPE}}}}$ & $\overline{\text{\textbf{\textit{RPE}}}}$ & & $\overline{\text{\textbf{\textit{RPE}}}}$ & $\overline{\text{\textbf{\textit{RPE}}}}$ & & $\overline{\text{\textbf{\textit{RPE}}}}$ & $\overline{\text{\textbf{\textit{RPE}}}}$ \\
				\textbf{\textit{Sequence}} & \textbf{\textit{Position}} & \textbf{\textit{Rotation}} & & \textbf{\textit{Position}} & \textbf{\textit{Rotation}} & & \textbf{\textit{Position}} & \textbf{\textit{Rotation}} \\
				& \textbf{\textit{($\times 10^{-3}$)}} & \textbf{\textit{($\times 10^{-3}$ deg/m)}} & & \textbf{\textit{($\times 10^{-3}$)}} & \textbf{\textit{($\times 10^{-3}$ deg/m)}} & & \textbf{\textit{($\times 10^{-3}$)}} & \textbf{\textit{($\times 10^{-3}$ deg/m)}} \\ 
				\hline
				shapes\_6dof & 50 & 114 & & 1 & 6 & & 2 & 6 \\ 
				shapes\_translation & 92 & 169 & & 1 & 3 & & 1 & 2 \\ 
				poster\_6dof & 18 & 57 & & 1 & 6 & & 2 & 7 \\ 
				poster\_translation & 5 & 6 & & 4 & 5 & & 4 & 5 \\ 
				hdr\_poster & 2 & 7 & & 3 & 7 & & 3 & 7 \\ 
				boxes\_6dof & 6 & 8 & & 1 & 3 & & 1 & 2 \\ 
				boxes\_translation & 4 & 4 & & 2 & 4 & & 2 & 4 \\ 
				hdr\_boxes & 4 & 6 & & 2 & 6 & & 2 & 6 \\ 
				dynamic\_6dof & 32 & 117 & & 2 & 5 & & 2 & 5 \\ 
				dynamic\_translation & 34 & 32 & & 4 & 5 & & 4 & 5 \\ 
				\hline
				\hline
			\end{tabular}
			\label{tab:ev:imu:results}
		\end{center}
	\end{table*}
	
	Table \ref{tab:ev:imu:results} compares different inertial configurations using the same RPE metrics of Table \ref{tab:ev:configs:results}. Both E-I-C2 and E-I-C3 configurations show a significant performance boost to the E-I-C1 configuration. In this case, the prolonged tracking period of E-I-C2 and E-I-C3  configurations provokes more IMU initialization and refinement steps and improves the accuracy of the IMU parameters and the whole system.
	
	The last two columns of Table \ref{tab:ev:imu:results} also investigate the efficacy of the adaptive selection of the tiny window size. Although the results are similar in most sequences, noting that the window length in E-I-C3 is fixed to the optimal value, a wrongly selected size can impact the performance of the system. Therefore, we report the E-I-C2 configuration as our preferred method.

	\begin{table*}
		\caption{Average translation (top) and yaw (bottom) errors for four distance ranges for sequences of \cite{ethz_pub_ds_Mueggler_2017}. 
			For each distance range, the values on the left of the column designate the results of the proposed algorithm (E-I-C2), compared against the results for event-inertial configuration in \cite{ultimate_slam_vidal_2018} on the right. Empty spaces (- symbol) show tracking is lost, and no value is available.}
		\ra{1.3}
		\begin{center}
			\begin{tabular}{@{}lccccccccccc@{}}
				\hline
				\hline
				\multicolumn{12}{c}{\textbf{Average Translation Error (m)}} \\
				\textbf{Sequence} & \multicolumn{11}{c}{\textbf{Distance (m)}} \\
				& \multicolumn{2}{c}{\textbf{10.0}} & & \multicolumn{2}{c}{\textbf{20.0}} & &  \multicolumn{2}{c}{\textbf{30.0}} & & \multicolumn{2}{c}{\textbf{40.0}} \\
				\hline
				shapes\_6dof & 0.15 & 0.31 & & 44.56 & 0.41 & & 
				-
				& 0.49 & & 
				- 
				& 0.55 \\
				shapes\_translation & 0.83 & 0.26 & & 7.60 & 0.79 & & 72.01 & 1.35 & & 
				- 
				& 1.66 \\ 
				poster\_6dof & 0.23 & 
				- 
				& & 1.43 & - & & 9.76 & - & & 20.47 & - \\
				poster\_translation & 0.12 & 0.09 & & 0.20 & 0.13 & & 0.30 & 0.27 & & 0.34 & 0.25 \\
				hdr\_poster & 0.04 & 0.19 & & 0.31 & 0.33 & & 0.38 & 0.45 & & 0.90 & 0.35 \\
				boxes\_6dof & 0.23 & 0.08 & & 0.31 & 0.28 & & 0.46 & 0.36 & & 1.03 & 0.48 \\
				boxes\_translation & 0.12 & 0.28 & & 0.24 & 0.26 & & 0.42 & 0.22 & & 0.53 & 0.38 \\
				hdr\_boxes & 0.13 & 0.17 & & 0.06 & 0.32 & & 0.15 & 0.48 & & 0.37 & 0.52 \\
				dynamic\_6dof & 0.16 & 0.10 & & 3.83 & 0.38 & & 27.98 & 0.42 & & - & - \\
				dynamic\_translation & 0.26 & 0.12 & & 1.25 & 0.56 & & - & 0.65 & & - & - \\
				\hline
			\end{tabular} \\ \vspace{2mm}
			\begin{tabular}{@{}lccccccccccc@{}}
				\hline
				\hline
				\multicolumn{12}{c}{\textbf{Average Yaw Error (deg)}} \\
				\textbf{Sequence} & \multicolumn{11}{c}{\textbf{Distance (m)}} \\
				& \multicolumn{2}{c}{\textbf{10.0}} & & \multicolumn{2}{c}{\textbf{20.0}} & &  \multicolumn{2}{c}{\textbf{30.0}} & & \multicolumn{2}{c}{\textbf{40.0}} \\
				\hline
				shapes\_6dof & 2.11 & 4.13 & & 10.18 & 2.66 & & 10.28 & 1.05 & & 22.87 & 0.94 \\
				shapes\_translation & 1.19 & 2.28 & & 3.53 & 5.21 & & 6.29 & 8.37 & & 7.13 & 9.16 \\ 
				poster\_6dof & 0.72 & 53.42 & & 1.25 & 32.59 & & 3.74 & 18.01 & & 2.43 & 40.46 \\
				poster\_translation & 0.32 & 0.38 & & 0.30 & 2.66 & & 0.53 & 4.89 & & 0.51 & 7.40 \\
				hdr\_poster & 2.09 & 3.35 & & 3.44 & 4.86 & & 3.91 & 2.94 & & 4.13 & 4.16 \\
				boxes\_6dof & 5.74 & 2.96 & & 2.35 & 2.05 & & 3.56 & 1.26 & & 3.60 & 2.19 \\
				boxes\_translation & 2.67 & 0.54 & & 4.02 & 1.78 & & 5.48 & 5.19 & & 6.07 & 3.96 \\
				hdr\_boxes & 2.66 & 0.62 & & 4.29 & 1.31 & & 4.78 & 1.72 & & 5.74 & 3.10 \\
				dynamic\_6dof & 3.17 & 3.37 & & 5.18 & 4.11 & & 4.01 & 5.82 & & - & - \\
				dynamic\_translation & 1.49 & 4.69 & & 4.19 & 1.10 & & - & 1.85 & & - & - \\
				\hline
			\end{tabular}
			\label{tab:ev:imu:comp:uslam}
		\end{center}
	\end{table*}
	
\begin{figure*}[htbp]
		\centering
		\begin{subfigure}{.31\textwidth}
			\centering
			\includegraphics[trim=15 10 45 35,clip,width=\linewidth]{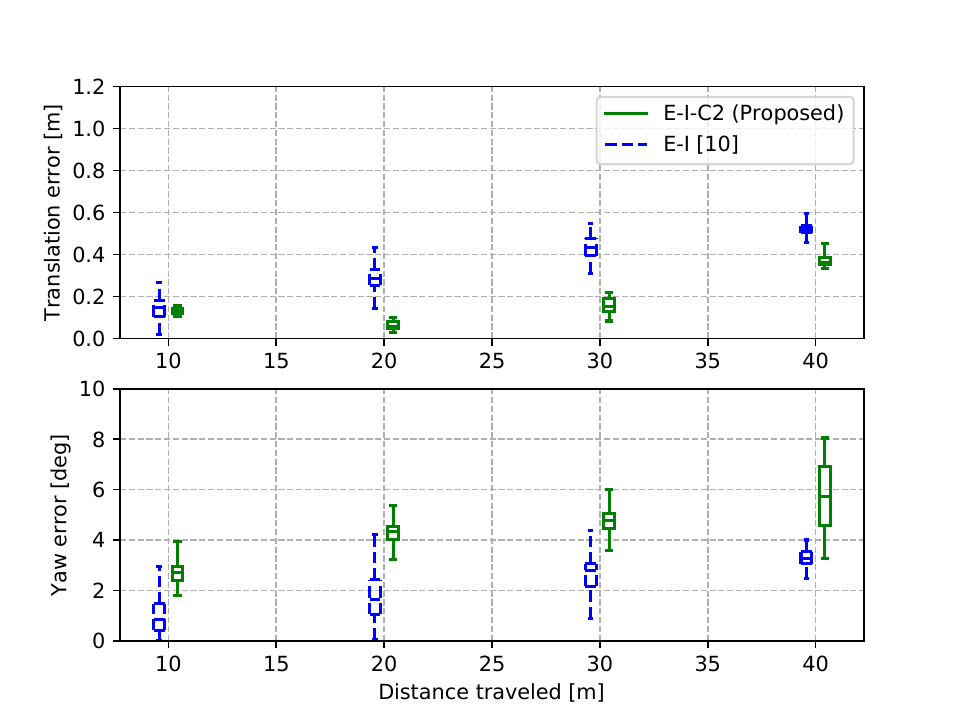}
			\caption{}
			\label{subfig:hdr:boxes}
		\end{subfigure}\quad
		\begin{subfigure}{.31\textwidth}
			\centering
			\includegraphics[trim=15 10 45 35,clip,width=\linewidth]{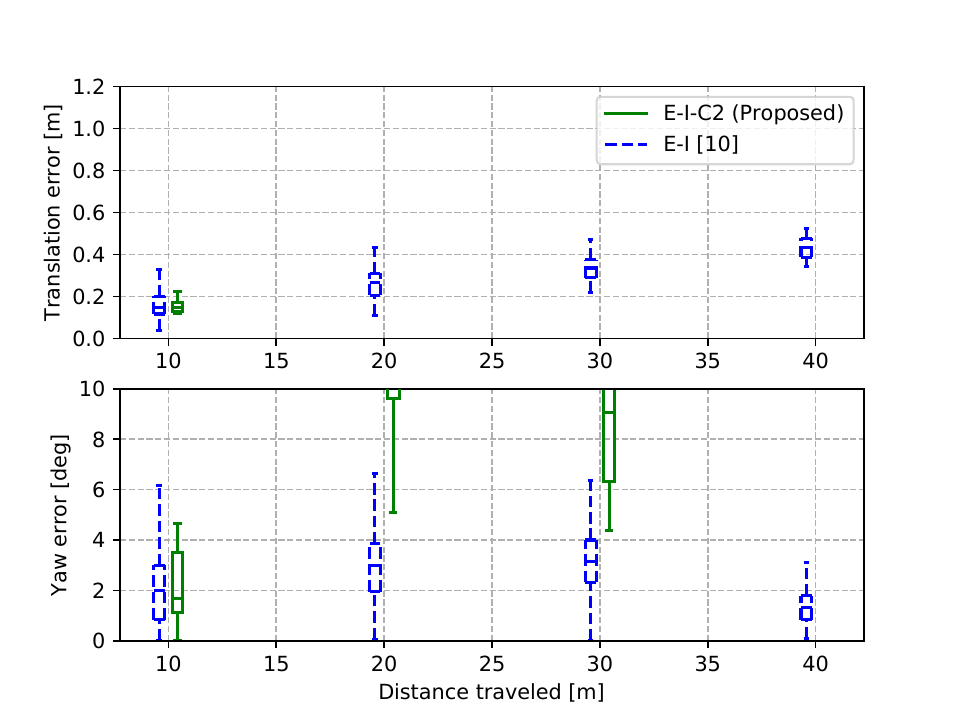}
			\caption{}
			\label{subfig:shapes:6dof}
		\end{subfigure}\quad
		\begin{subfigure}{0.31\textwidth}
			\centering
			\includegraphics[trim=15 10 45 35,clip,width=\linewidth]{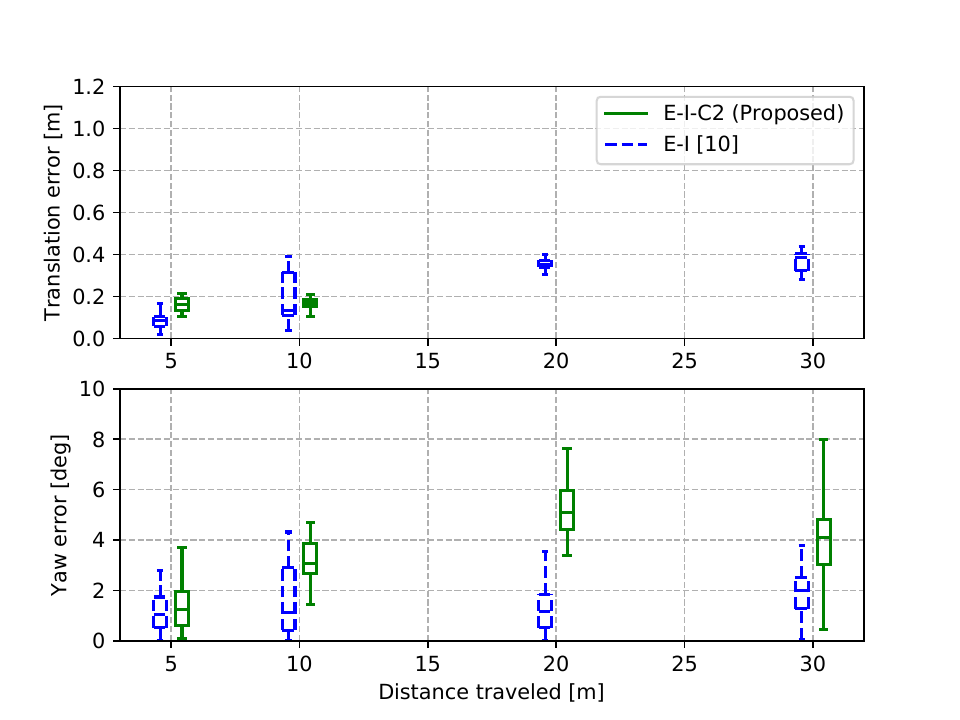}
			\caption{}
			\label{subfig:dynamic:6dof}
		\end{subfigure}
		\caption{Comparison between the proposed event-inertial SLAM pipeline and the the method described in \cite{ultimate_slam_vidal_2018}.
			(\subref{subfig:hdr:boxes}) hdr\_boxes, 
			(\subref{subfig:shapes:6dof}) shapes\_6dof, 
			(\subref{subfig:dynamic:6dof}) dynamic\_6dof.}
		\label{fig:comp:uslam}
	\end{figure*}

	Next, we compare our pipeline against the most relevant algorithm that provides a stable open-source implementation.
	Table \ref{tab:ev:imu:comp:uslam} and Fig. \ref{fig:comp:uslam} contrast the E-I-C2 configuration of the proposed algorithm against the event-inertial method in \cite{ultimate_slam_vidal_2018}. We run the latest open-source release of their algorithm%
	\footnote{https://github.com/uzh-rpg/rpg\_ultimate\_slam\_open} 
	for each sequence in \cite{ethz_pub_ds_Mueggler_2017} using the default configurations (the sequence poster\_6dof is excluded because this implementation of \cite{ultimate_slam_vidal_2018} fails to produce reliable results). Similar to \cite{ultimate_slam_vidal_2018}, we measure and report the average relative translation and yaw errors over a range of distances. We calculate the relative pose error between the first one hundred pairs that meet a specific distance range.
	
	Although the window-based state estimator of \cite{ultimate_slam_vidal_2018} estimates the pose of the camera and map points simultaneously, our algorithm relies on an accurate map to find the next pose. As a result, any condition that degrades map estimate accuracy will affect our results. For the sequence shapes\_6dof, when the camera moves fast in front of a low-textured poster, the tracking fails due to the lack of detected features, and errors grow exponentially. Even though the scene in sequence dynamic\_6dof is textured, the change in 3D feature locations affects the performance of the proposed method. Despite the poor lighting condition in hdr\_boxes, there are enough reliable map points, and the proposed algorithm outperforms \cite{ultimate_slam_vidal_2018} based on the translation errors. Since the reliability of the map estimate is an integral part of our method, we ensure its accuracy based on several heuristics, such as the number of successfully tracked map points and the reprojection error.
	
	\begin{figure}[htbp]
		\centering
		\includegraphics[trim=240 75 180 50,clip,width=.45\textwidth]{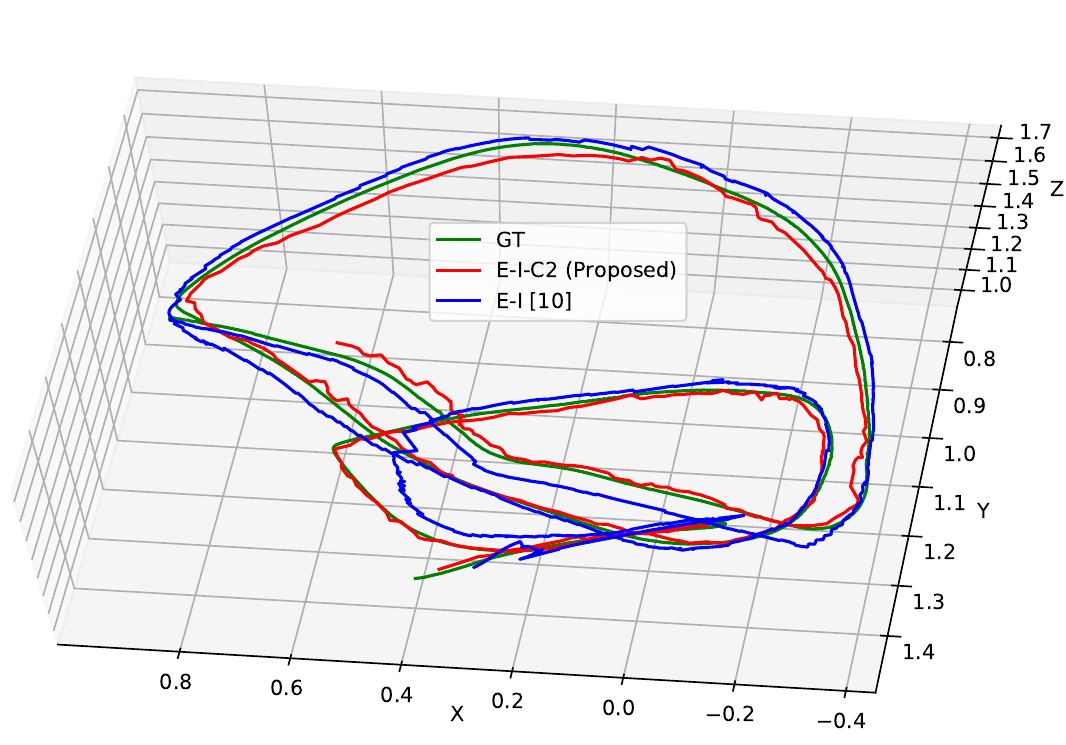}
		\caption{%
			The absolute estimated positions of the proposed algorithm and the event-inertial method of \cite{ultimate_slam_vidal_2018} are compared against the ground truth in terms of the APE for the first 15 seconds of the sequence boxes\_6dof.
		}
		\label{fig:compPoseEstGT}
	\end{figure}
	
	Fig. \ref{fig:compPoseEstGT} compares the absolute estimated position of the results of Fig. \ref{fig:comp:uslam} for the sequence boxes\_6dof. We use the APE tools in \cite{orb_slam_Mur_Artal_2015} to align the first 15 seconds of the estimates with the ground truth. In this case, the proposed method can track the ground truth more accurately.
	
	As discussed in Section \ref{sec:related:work}, MC-VEO \cite{mc_veo_2024} uses both the events and intensity images to estimate the 6DoF pose of a DAVIS device. Since the motion compensation method of this work is the most relevant to ours, we evaluate the absolute translation error discussed in \cite{mc_veo_2024} for the event-only version of our pipeline for four sequences of \cite{stereo_davis_dataset_2018} and provide our findings here for reference. Our results are summarized in Table \ref{tab:ev:mcveo:results}, along with that of \cite{mc_veo_2024}, which are derived from their work. As expected, the use of both events and images for this specific dataset produces superior results. However, we argue that the MCI generation module of our event-only approach can produce sufficiently accurate results, which can prove effective in situations where the quality of images is severely degraded, e.g., navigation at night.
	
	\begin{table}
		\caption{The absolute translation error reported for the proposed event-only method and the algorithm described in \cite{mc_veo_2024} for four sequences of the Stereo DAVIS Dataset \cite{stereo_davis_dataset_2018} (figures for the latter are derived from the original paper).}
		\ra{1.0}
		\begin{center}
			\begin{tabular}{@{}lcc@{}}
				\hline
				\hline & & \\[-1.5ex]
				& \multicolumn{2}{c}{\textbf{ATE (cm)}} \\
				\cmidrule{2-3}
				\textbf{\textit{Sequence}} & \textbf{Event-only (ours)} & \textbf{E-F} \cite{mc_veo_2024} \\
				\hline & & \\[-1.5ex]
				bin & 6.487 & 1.138 \\ 
				boxes & 8.541 & 2.085 \\ 
				desk & 6.936 & 1.584 \\ 
				monitor & 1.657 & 0.953 \\ 
				\hline
				\hline
			\end{tabular}
			\label{tab:ev:mcveo:results}
		\end{center}
	\end{table}

	Although we have not tried to optimize the performance of the proposed algorithm, we present the average tracking time statistics to contrast the computation cost of each module. In particular, we measure and report the performance of the main tracking thread (MTH), the first-level event image reconstruction module (L1), and the second-level event frame tracking thread (L2). Since it almost takes several tiny frames in L1 and one iteration of L2 for the main tracking thread to process input events, we include the timing as per MCI and as per tiny frame (TF). Assuming there are three TF for each MCI on average, we subtract double L1 values from the MCI results and report the per TF results. The timing statistics for the local mapping thread are not reported here. We run our pipeline on a system with an Intel Core-i7 9700K CPU, 64 GB RAM, and Ubuntu 20.04 LTS operating system.
	
	Table \ref{tab:timing:results} summarizes the timing results of different components of our pipeline for two sensor configurations and some representative sequences of both datasets \cite{ethz_pub_ds_Mueggler_2017} and \cite{mvsec_ds_Zhu_2018}. The overall performance of our algorithm is several times lower than the real-time. Note that more textured sequences demand more processing power. The addition of the IMU slightly increases the performance cost of our method. Based on these results, the reconstruction of each MCI is generally the most costly operation. It takes about the same amount of time to reconstruct an MCI, around 28 ms on average, consistent across all configurations and datasets. Although we acknowledge that the latency introduced with MCI reconstruction can impact the advantages of using low-latency event data, these effects can be mitigated using motion priors from L1 or IMU measurements between these intervals.
	
	The timing results of L2 are only comparable across the Public Event dataset. For sequences of \cite{mvsec_ds_Zhu_2018}, the cost of the inertial L2 module is almost twice the event-only case. One reason for this difference could be the high depth variation inherent in the fisheye camera and the difficulty of scale estimation and refinement involved in inertial tracking in this case (although the local mapping is performed in a separate thread, the inertial L2 must wait because of the map change).
	
	\begin{table*}
		\caption{Average tracking time statistics in milliseconds for three main modules of the proposed algorithm, calculated for representative sequences of \cite{ethz_pub_ds_Mueggler_2017} and \cite{mvsec_ds_Zhu_2018}.}
		\ra{1.3}
		\begin{center}
			\begin{tabular}{@{}lccccccccc@{}}
				\hline
				& \multicolumn{4}{c}{\textbf{Event-Only}} & & \multicolumn{4}{c}{\textbf{Event-Inertial-C1}} \\
				\cmidrule{2-5} \cmidrule{7-10} 
				& \textbf{\textit{MTH}} & \textbf{\textit{MTH}} & \textbf{\textit{L1}} & \textbf{\textit{L2}} & & \textbf{\textit{MTH}} & \textbf{\textit{MTH}} & \textbf{\textit{L1}} & \textbf{\textit{L2}} \\
				\textbf{\textit{Sequence}} & \textbf{\textit{/MCI}} & \textbf{\textit{/TF}} & \textbf{\textit{/TF}} & \textbf{\textit{/TF}} & & \textbf{\textit{/MCI}} & \textbf{\textit{/TF}} & \textbf{\textit{/TF}} & \textbf{\textit{/TF}} \\
				\hline
				shapes\_6dof & 72 & 48 & 12 & 6 & & 82 & 56 & 13 & 7 \\ 
				poster\_6dof & 158 & 106 & 26 & 7 & & 173 & 119 & 27 & 9 \\ 
				boxes\_6dof & 158 & 106 & 26 & 7 & & 174 & 118 & 28 & 9 \\
				dynamic\_6dof & 138 & 92 & 23 & 7 & & 157 & 107 & 25 & 9 \\
				\hline
				indoor\_flying1 & 139 & 93 & 23 & 8 & & 147 & 99 & 24 & 15 \\ 
				outdoor\_day1 & 171 & 115 & 28 & 5 & & 185 & 127 & 29 & 10 \\ 
				outdoor\_night1 & 187 & 125 & 31 & 5 & & 202 & 138 & 32 & 14 \\ 
				\hline
			\end{tabular}
			\label{tab:timing:results}
		\end{center}
	\end{table*}

	\section{Conclusion} \label{sec:conclusion}
	
	We proposed an algorithm that reconstructs MCIs from the adaptively selected event windows and uses them to estimate the structure and trajectory. The first module tracks FAST features across multiple tiny frames to choose the best event window size and resolve the parameters needed for MCI generation. After reconstruction, it sends the best MCI representation to an image-based SLAM to initialize the map and track the 6-DOF pose of the camera in it. Furthermore, we showed how to utilize inertial measurements to improve the performance of the event-only algorithm.
	
	We compared the estimated trajectory of different configurations of the proposed pipeline with the ground truth for sequences of two publicly available event datasets. Although the algorithm produces accurate results in most cases, its performance depends on the reliability of the map estimate. We also showed that the proposed algorithm outperforms similar event-inertial methods as long as a valid map is available.
	
	Although the event window selection module is costly, it can still be used in most event-based algorithms to select the best event window without reconstructing the MCI. It is possible to use different techniques such as relocalization or loop-closing to improve the accuracy of the map, the overall performance, and the algorithm's stability. These techniques generally extract and use descriptors to link different segments of a pose graph. One can consider event-based descriptors for that purpose or extract descriptors from regular intensity images, which are processed along with the event stream.

	\bibliographystyle{ieeetr}
	\bibliography{References} 
	
\end{document}